\documentclass[letterpaper]{article} 
\usepackage[]{aaai25}  
\usepackage{times}  
\usepackage{helvet}  
\usepackage{courier}  
\usepackage[hyphens]{url}  
\usepackage{graphicx} 
\usepackage{natbib}  
\usepackage{caption} 
\usepackage{algorithm}
\usepackage{algorithmic}

\usepackage{amsmath,amssymb,amsfonts}
\usepackage{algorithmic}
\usepackage{graphicx}
\usepackage{textcomp}
\usepackage{algorithm,algorithmic}
\usepackage{multirow}
\usepackage{newfloat}
\usepackage{listings}
\DeclareCaptionStyle{ruled}{labelfont=normalfont,labelsep=colon,strut=off} 
\floatstyle{ruled}
\newfloat{listing}{tb}{lst}{}
\floatname{listing}{Listing}
\title{Synergy-Guided Regional Supervision of Pseudo Labels for Semi-Supervised Medical Image Segmentation}
\author {
    Tao Wang\textsuperscript{\rm 1},
    Xinlin Zhang\textsuperscript{\rm 1},
   Yuanbin Chen\textsuperscript{\rm 1},
   Yuanbo Zhou\textsuperscript{\rm 1},
   Longxuan Zhao\textsuperscript{\rm 1},
   Tao Tan\textsuperscript{\rm 2},
   Tong Tong\textsuperscript{\rm 1}
}
\affiliations {
    \textsuperscript{\rm 1}College of physics and information engineering, University of Fuzhou University, Fuzhou 350108, China\\
    \textsuperscript{\rm 2}Faculty of Applied Science, University of Macao Polytechnic University, Macao\\
}

\usepackage{bibentry}

\begin{document}

\maketitle

\begin{abstract}
Semi-supervised learning has received considerable attention for its potential to leverage abundant unlabeled data to enhance model robustness. Pseudo labeling is a widely used strategy in semi supervised learning. However, existing methods often suffer from noise contamination, which can undermine model performance. To tackle this challenge, we introduce a novel Synergy-Guided Regional Supervision of Pseudo Labels (SGRS-Net) framework. Built upon the mean teacher network, we employ a Mix Augmentation module to enhance the unlabeled data. By evaluating the synergy before and after augmentation, we strategically partition the pseudo labels into distinct regions. Additionally, we introduce a Region Loss Evaluation module to assess the loss across each delineated area. Extensive experiments conducted on the LA dataset have demonstrated superior performance over state-of-the-art techniques, underscoring the efficiency and practicality of our framework. The code is available in the Supplementary Material.
\end{abstract}

%

\section{Introduction}
With the rapid development of computer vision and deep learning, automatic segmentation of medical images has become a focal point of research. Supervised learning segmentation methods, such as U-Net \cite{unet} and UNeXt \cite{unext} have achieved significant success. However, traditional medical image segmentation methods often rely on a large amount of precisely annotated training data, which is expensive and time-consuming to obtain. This dependency limits the scalability and applicability of these methods. In recent years, semi-supervised learning has attracted widespread attention for its potential to utilize fewer labeled data in conjunction with abundant unlabeled data, thereby enhancing model generalization. Numerous semi-supervised medical image analysis methods have been introduced, including pseudo labels \cite{CHAITANYA2023102792}, deep co-training \cite{ZHENG2022106051}, deep adversarial learning \cite{9966841}, the mean teacher and its extensions \cite{NIPS2017_68053af2}\cite{li2020transformation}, and contrastive learning \cite{WANG2022102447}, among others. These methods effectively leverage both labeled and unlabeled data to develop robust models.

Pseudo-label learning is widely adopted. This method begins by training the model on labeled data, then generating pseudo labels for unlabeled data using predicted probability maps. These pseudo labels are combined with labeled data for further model training, enhancing accuracy and generalization \cite{lee2013pseudo}. Building on this approach, Lu et al. \cite{LU2023126411} introduced two auxiliary decoders into a network to generate pseudo labels from these auxiliary decoders. Similarly, Li et al. \cite{slupl} explored generating pseudo labels through the cyclic optimization of neural networks, incorporating self-supervised tasks. Additionally, Luo et al. \cite{crossteaching} developed the Cross Teaching framework, where the prediction of one network serves as the pseudo label to directly guide another network in an end-to-end manner.

While these methods effectively enhance model performance, the inherent inaccuracy of pseudo labels compared to ground truth remains a challenge. While some pseudo labels are comparable to ground truth, others may be compromised by noise. As iterative training progresses, this noise can adversely affect models robustness \cite{9729424}. Although applying noise-robust loss functions can mitigate the effects of noise to some extent \cite{pmlr-v151-olmin22a}, optimizing against unnecessary noise suppression for high-quality pseudo labels may reduce model accuracy due to over-smoothing \cite{tian2023learning}.

To effectively utilize pseudo labels, we introduce the Synergy-Guided Regional Supervision (SGRS-Net). During training, a Pseudo Label Generation (PLG) Module is employed within the mean-teacher framework and transform the predictions of the teacher network into pseudo labels. Additionally, we introduce a Mix Augumentation (MA) Module which utilize annotated data to augment unlabeled data. Following this, our Synergy Evaluation module partitions pseudo labels into different regions and then apply distinct loss functions to evaluate the corresponding loss for each region. This approach is designed to maximize the use of pseudo labels while minimizing the impact of potential noise. In summary, our contributions are listed as follows:

\begin{itemize}
	\item We introduce a novel SGRS-Net framework for semi-supervised medical image segmentation. This framework employs a PLG module to generate pseudo labels and a MA module to enhance the diversity of the unlabeled dataset. 
	\item Based on PLG and MA modules, we propose a Synergy Evaluation module and a Regional Loss Evaluation module to mitigate the impact of noise while fully leveraging the supervisory signal from pseudo labels.
	\item SGRS-Net was evaluated on three public available datasets, showing promising results, especially when using only 5$\%$ of labeled data. This highlights the model's effectiveness in scenarios with limited labeled data.
\end{itemize}
\vspace{-0.3cm}
\section{Relate Work}
\subsection{Medical Image Segmentation}
Deep learning has demonstrated promising performance in various image segmentation tasks. In medical image segmentation, U-Net \cite{unet} and its variants have become benchmark methods for further research and practical applications\cite{8932614}. 
The introduction of V-Net \cite{milletari2016v} and 3D-Unet \cite{10.1007/978-3-319-46723-8_49} extended U-Net to 3D medical image segmentation. Subsequently, attention mechanisms were introduced to enhance feature representation in the channel and spatial dimensions. Attention UNet \cite{schlemper2019attention} integrated attention gates into U-Net to refine skip-connected low-level features. Additionally, many high-performance algorithms have been developed such as CE-Net \cite{8662594} and UNeXt \cite{unext}. Although these methods have achieved promising performance, they are typically designed for fully supervised settings and their effectivenesses are limited by the costly labeled data.
\vspace{-0.2cm}
\subsection{Semi-supervised Medical Image Segmentation}
Many efforts have been made in semi-supervised medical image segmentation. Wu et al. \cite{wu2022mutual} proposed MC-Net, featuring a shared encoder and multiple slightly different decoders. The model is trained by regularizing the differences between decoder outputs. Peiris et al. \cite{peiris2023uncertainty} proposed a dual-view framework based on adversarial learning for volumetric image segmentation. Bai et al. introduced BCP \cite{Bai_2023_CVPR}, demonstrating excellent performance by bidirectionally copy-pasting labeled and unlabeled data in a Mean Teacher architecture. 

Pseudo-labeling is also widely used in semi-supervised learning. Early studies \cite{lee2013pseudo}\cite{fan2020inf} trained models on labeled data and generated pseudo labels for unlabeled data, which were then used for iterative training. To better utilize pseudo labels, researchers have developed various methods, such as identifying reliable pseudo labels through uncertainty estimation \cite{sedai2019uncertainty}, improving pseudo labels with Conditional Random Fields (CRF) \cite{krahenbuhl2011efficient}.
Additionally, Qiu et al. \cite{qiu2023federated} proposed a federated pseudo-labeling strategy which leveraging embedded knowledge from labeled clients to generate pseudo labels for unlabeled clients. Peng et al.\cite{peng2021self} employed Self-Paced Contrastive Learning to select confident pseudo labels for self-training. Luo et al. \cite{crossteaching} proposed a co-training method where two models are trained simultaneously, using their predictions provide pseudo-supervised signals to the other. Wang et al. \cite{wang2023mcf} introduced MCF, a dynamic competitive pseudo-label generation method, which contains two subnetworks. By evaluating performance between subnetworks in real-time, more reliable pseudo labels are dynamically selected during training.

Although these methods have made progress in improving the effectiveness of pseudo labels, they are still susceptible to noise.

\vspace{-0.2cm}
\section{Method}
This study introduces the SGRS-Net for semi-supervised medical image segmentation. For convenience, the notions and notations used in this paper are summarized in Table \ref{table1}. 
\vspace{-0.2cm}
\begin{table}[htbp]	
	\caption{Descriptions about key natations}
	\label{table1}
	\vspace{-0.2cm}
	\resizebox{0.48\textwidth}{!}{
		\begin{tabular}{ll}
			\hline
			\multicolumn{1}{l}{\makebox[0.1\textwidth][l]{Notations}} & Descriptions        \\ \hline
			$\mathcal{D}_{L}$, $\mathcal{GT}$                  & Labeled Image and correspond Ground Truth \\
			$\mathcal{D_{U}}$, $\mathcal{Y}$ 							  & Unlabeled Image and correspond Pseudo Label 	 \\
			$\mathcal{D_{M}}$ 							  & Image mixed from $\mathcal{D_{U}}$ and $\mathcal{D}_{L}$ by the MA module 	 \\
			$\theta_S$,$\theta_T$						  & Parameters of the student and teacher network	 \\
			$\mathrm{\Delta}$, $\mathrm{\Omega}$, $\mathrm{\Theta}$						      & Regions regard as disregarded, consistent, and inconsistent	 \\
			\hline
	\end{tabular} }
\end{table}
\vspace{-0.8cm}
\subsection{Overall architecture design}
\begin{figure*}[]
	\centering
	\includegraphics[width=1\textwidth]{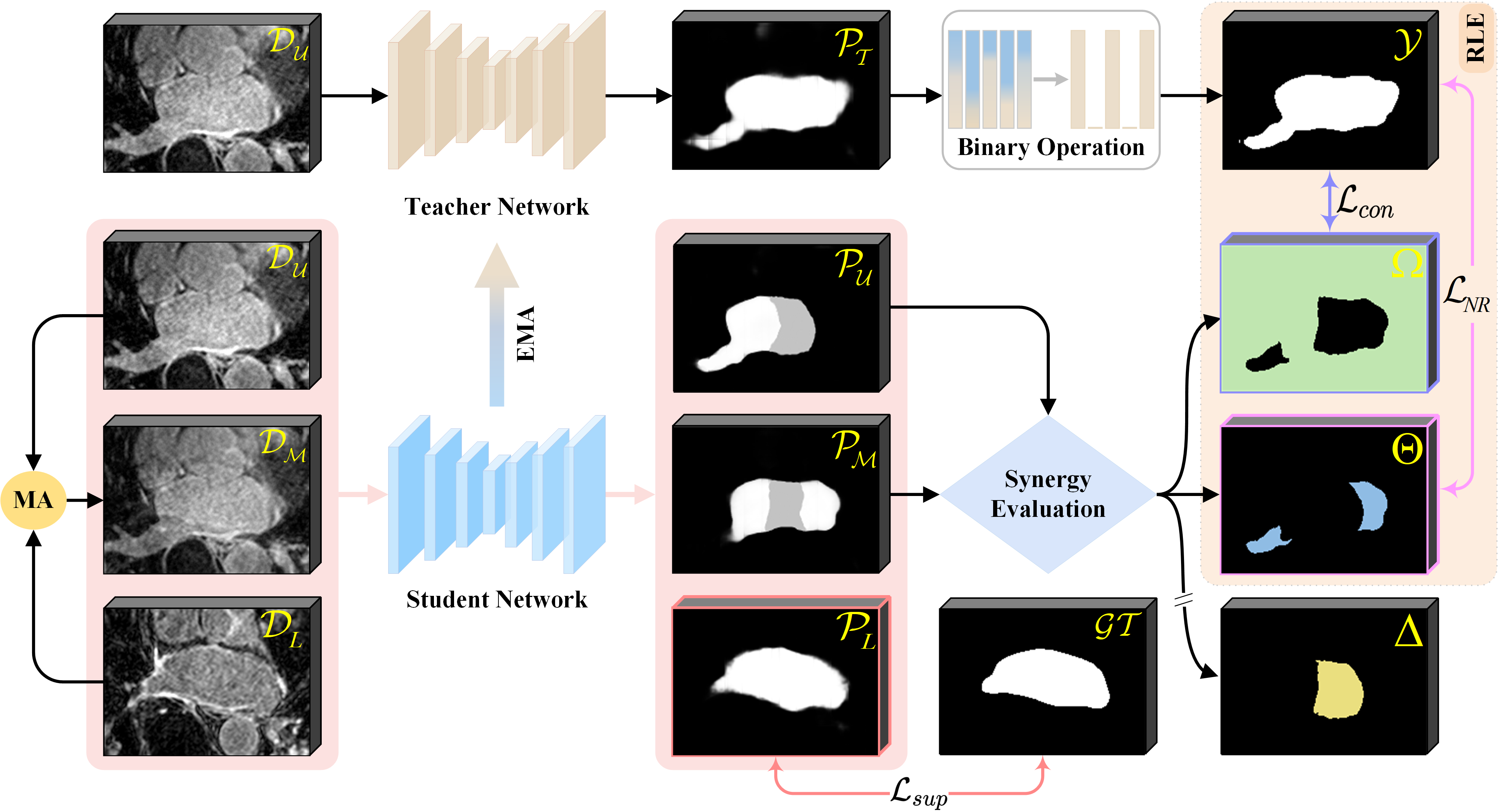}
	\caption{Overview of our proposed SGRS-Net. The teacher network updates its parameters from the student network using exponential moving averages (EMA). The "MA" denotes the Mix Augmentation Module, and the "RLE" represents the Regional Loss Evaluation module. }
	\label{figure1}
	\vspace{-0.6cm}
\end{figure*}

Figure. \ref{figure1} illustrates the framework of the proposed SGRS-Net. During training, we start by assessing unlabeled data with the teacher network to obtain $\mathcal{Y}$. Subsequently, the MA module enhances $\mathcal{D_{U}}$, resulting in $\mathcal{D_{M}}$. The student network processes $\mathcal{D}_{L}$, $\mathcal{D_{U}}$, and $\mathcal{D_{M}}$, producing the corresponding predictions $\mathcal{P}_{L}$, $\mathcal{P_{U}}$ and $\mathcal{P_{M}}$. Next, the Synergy Evaluation (SE) module evaluates the synergy between $\mathcal{P_{U}}$ and $\mathcal{P_{M}}$, and partition $\mathcal{Y}$ into three regions: $\mathrm{\Delta}$, $\mathrm{\Omega}$, and $\mathrm{\Theta}$. Finally, the Regional Loss Evaluation (RLE) module is used to evaluate the losses. Detailed explanations of each module will be provided in the subsequent sections.
\vspace{-0.2cm}
\subsection{Pseudo Label Generation Module}
To obtain pseudo labels for $\mathcal{D_{U}}$, different from the traditional method that rely on a single backbone network, we adopt a mean teacher framework. The teacher network evaluates the $\mathcal{D_{U}}$ to generate the corresponding $\mathcal{Y}$. The specific formula is presented as follows:  
\begin{equation}
	\mathcal{P_{T}} = f_{\theta_{T}}(\mathcal{D_{U}})
\end{equation}
\begin{equation}
	\mathcal{Y} = \mathrm{ArgMax}(\mathcal{P_{T}})
\end{equation}
where $f_{\theta_{T}}$ represents the processing by the teacher network.
\vspace{-0.2cm}
\subsection{Mix Augmentation Module}
To enhance the diversity of the $\mathcal{D_{U}}$, we introduce a data augmentation strategy inspired by the commonly employed Mix-Up method in visual tasks \cite{zhang2017mixup}. For a pair of samples $\mathcal{D_{U}}$ and $\mathcal{D}_{L}$, we perform a random proportion of linear interpolation to generate $\mathcal{D_{M}}$ for model training. The specific operations are formulated as follows:
\begin{equation}
	\lambda = max[Beta(\alpha ,\alpha),1-Beta(\alpha ,\alpha)]
\end{equation}
\begin{equation}
	\mathcal{D_{M}} = \lambda \cdot \mathcal{D_{U}} + (1 - \lambda) \cdot \mathcal{D}_{L}
\end{equation}
where $\alpha$ is a randomly generated hyperparameter. Notably, our method primarily focuses on $\mathcal{D_{U}}$, incorporating $\mathcal{D}_{L}$ to enhance the diversity of the unlabeled data without using $\mathcal{GT}$ or adding additional annotation information. Therefore, we use the $\mathcal{Y}$ corresponding to $\mathcal{D_{U}}$ as the pseudo label for $\mathcal{D_{M}}$ in subsequent steps. Additionally, the MA module serves as the basis for the consistency evaluation conducted by the SE modules.

\subsection{Synergy Evaluation Module}
To enhance the utilization of valuable information within pseudo labels and reduce the impact of noise, we introduce a SE module. Unlike traditional methods that typically focus on regularizing the consistency between predictions during training, we segments the $\mathcal{Y}$ into three parts: $\mathrm{\Delta}$, $\mathrm{\Theta}$, and $\mathrm{\Omega}$, based on their consistency. Specifically, we begin by assessing the information entropy of $\mathcal{P_{U}}$ and $\mathcal{P_{M}}$. Since higher entropy within the prediction signals greater uncertainty, we introduce a threshold, $\tau$. Regions with entropy exceed $\tau$ are considered unreliable and categorized as $\mathrm{\Delta}$. The procedures for this categorization are detailed as follows:
\begin{equation}
	\mathcal{P_{U}} ,\mathcal{P_{M}} = \mathrm{Softmax}(f_{\theta _{S}}(\mathcal{D_{U}})) , \mathrm{Softmax}(f_{\theta _{S}}(\mathcal{D_{M}}))
\end{equation}
\begin{equation}
	\mathrm{Ent}(p) = -\sum_{j\in C} p_{j} \cdot \log p_{j}
\end{equation}
\begin{equation}
	\mathrm{\Delta}  = (\mathrm{Ent}(\mathcal{P_{U}}) > \tau )  \lor  (\mathrm{Ent}(\mathcal{P_{M}})> \tau)
\end{equation}
Where $C$ represents the total number of categories and $p_{j}$ denotes the probability distribution of pixels for the $j$-th category. Regions that are clearly distinguishable with consistent results are labeled as $\mathrm{\Omega}$, while those with inconsistent categories are labeled as $\mathrm{\Theta}$. The specific operations are formulated as follows:
\begin{equation}
	\mathcal{A_{U}},\mathcal{A_{M}}= \mathrm{ArgMax}(\mathcal{P_{U}}), \mathrm{ArgMax}(\mathcal{P_{M}})
\end{equation}
\begin{equation}
	\mathrm{\Omega},\mathrm{\Theta} = (\mathcal{A_{U}} \odot \mathcal{A_{M}}) \land ( \neg\mathrm{\Delta}), (\mathcal{A_{U}} \oplus \mathcal{A_{M}}) \land ( \neg\mathrm{\Delta} )
\end{equation}

\subsection{Regional Loss Evaluation Module}
After assessing synergy, the $\mathcal{Y}$ is divided into three segments: $\mathrm{\Delta}, \mathrm{\Omega}, \mathrm{\Theta}$. Subsequently, we conduct an individual loss evaluation for each category. Regions classified as $\mathrm{\Delta}$ exhibit higher uncertainty, indicating that the model cannot provide a clear judgment for these areas. As a result, the corresponding pseudo labels are more likely to contain noise. To minimize the impact of noise, these regions are excluded from loss evaluation.
\vspace{-0.2cm}
\subsubsection{Learning from $\mathcal{GT}$}
During training, the $\mathcal{GT}$ serves as the gold standard for $\mathcal{D}_{L}$, and the supervised loss is evaluated using $\mathcal{L}_{sup}$. Based on the Cross Entropy Loss and Dice Loss, the specific function is defined as follows:
\begin{equation}
	\mathcal{L}_{pce}^{\mathbb{U}}(p,y) = - \sum_{i \in \mathbb{U}} \sum_{j\in C}y[i,j] \cdot \log p[i,j]
	\label{equation9}
\end{equation}	
\begin{equation}
	\mathcal{L}_{pdc}^{\mathbb{U}}(p,y) = 1-\frac{2 \times { \sum\limits_{i \in \mathbb{U}}{\sum\limits_{j \in C}}  p[i,j]\cdot y[i,j]} }{\sum\limits_{i \in \mathbb{U}}{ \sum\limits_{j \in C}}p[i,j]^{2}+\sum\limits_{i \in \mathbb{U}}{ \sum\limits_{j \in C}} y[i,j]^{2}}  
	\label{equation10}
\end{equation}
\begin{equation}
	\mathcal{L}_{sup} = \mathcal{L}_{pce}^{\mathbb{U}}(\mathcal{P}_{L},\mathcal{GT}) + \mathcal{L}_{pdc}^{\mathbb{U}}(\mathcal{P}_{L},\mathcal{GT})
	\label{equation11}
\end{equation}
where $\mathbb{U}$ represents all the pixels within the $\mathcal{GT}$. The $p[i,j]$ and $y[i,j]$ denote the predicted probability and the ground truth label for the pixel $i$ with class $j$, respectively.
\vspace{-0.2cm}
\subsubsection{Learning from $\mathrm{\Omega}$}
Regions identified as $\mathrm{\Omega}$, characterized with greater certainty and consistent predictions. Therefore, we believe that the model demonstrates strong judgment, and the region $\mathrm{\Omega}$ within $\mathcal{Y}$ can be trusted. We introduce a local consistency supervision loss $\mathcal{L}_{con}$, which learns from unlabeled data by minimizing the difference between the prediction and the $\mathcal{Y}$ within $\mathrm{\Omega}$. The loss is implemented using Cross Entropy Loss and Dice Loss. Based on equations \eqref{equation9} and \eqref{equation10}, the $\mathcal{L}_{con}$ is formulated as:
\begin{equation}
	\mathcal{L}_{c}^{\mathrm{\Omega}}(p,y) = \mathcal{L}_{pce}^{\mathrm{\Omega}}(p,y) +  \mathcal{L}_{pdc}^{\mathrm{\Omega}}(p,y) 
	\label{equation12}
\end{equation}
\begin{equation}
	\mathcal{L}_{con} = \mathcal{L}_{c}^{\mathrm{\Omega}}(\mathcal{P_{U}},\mathcal{Y}) + \mathcal{L}_{c}^{\mathrm{\Omega}}(\mathcal{P_{M}},\mathcal{Y}) 
	\label{equation13}
\end{equation}
\vspace{-0.2cm}
\subsubsection{Learning from $\mathrm{\Theta}$}
For regions within $\mathrm{\Theta}$, which display lower entropy indicating a degree of stability, yet show discrepancies in predicted categories. This observation suggests that the corresponding $\mathcal{Y}$ within this region may contain noise. To address this issue, we propose a regional noise-robust loss function, denoted as $\mathcal{L}_{NR}$, designed to mitigate the influence of noise within $\mathrm{\Theta}$. The formulation of $\mathcal{L}_{NR}$ is presented as follows:
\vspace{-0.2cm}
\begin{equation}
	\begin{split}
		\mathcal{L}_{psce}^{\mathrm{\Theta}}(p, y) = -\sum_{i \in \mathrm{\Theta}} \sum_{j\in C} [ (1 - \varepsilon ) \cdot y[i,j] \cdot \log(p[i,j]) +  \\ \frac{\varepsilon }{C} \log(p[i,j]) ]
	\end{split}
\end{equation}
\vspace{-0.2cm}
\begin{equation}
	\mathcal{L}_{psdc}^{\mathrm{\Theta}}(p,y) = 1-\frac{2 \times { \sum\limits_{i \in \mathrm{\Theta}}{ \sum\limits_{j \in C}}  p[i,j]\cdot y[i,j]} }{\sum\limits_{i \in \mathrm{\Theta}}{ \sum\limits_{j \in C}}p[i,j]^{2}+\sum\limits_{i \in \mathrm{\Theta}}{ \sum\limits_{j \in C}} y[i,j]^{2} + \eta }  
\end{equation}
\begin{equation}
	\mathcal{L}_{s}^{\mathrm{\Theta}}(p,y) = \mathcal{L}_{psce}^{\mathrm{\Theta}}(p,y) +  \mathcal{L}_{psdc}^{\mathrm{\Theta}}(p,y) 
\end{equation}
\begin{equation}
	\mathcal{L}_{NR} = \mathcal{L}_{s}^{\mathrm{\Theta}}(\mathcal{P_{U}},\mathcal{Y}) + \mathcal{L}_{s}^{\mathrm{\Theta}}(\mathcal{P_{M}},\mathcal{Y})
	\label{equation17}
\end{equation}
where $\varepsilon$ and $\eta$ are the smoothing parameter.
\vspace{-0.2cm}
\subsection{Total Loss Funciton}
The proposed SGRS-Net framework aims to learn from both labeled and unlabeled data, the total loss is articulated as:
\begin{equation}
	\mathcal{L} = \mathcal{L}_{sup} + \lambda \cdot (\mathcal{L}_{con}+\mathcal{L}_{NR})
\end{equation}
where $\mathcal{L}_{sup}$, $\mathcal{L}_{con}$ and $\mathcal{L}_{NR}$ are already illustrated in equations \eqref{equation11}, \eqref{equation13} and \eqref{equation17}, respectively. We incorporate $\lambda(t)$, a widely utilized time-dependent Gaussian warming-up function \cite{laine2016temporal}, to modulate the balance between losses at different training stages. It is defined as:
\begin{align}
	\lambda (t) & = \left\{\begin{matrix}e^{(-5(1-\frac{t}{t_{warm}} )^2)} \quad t < t_{warm} \\
		\hspace*{0.0em}1 \hspace*{6.2em} t \ge t_{warm}
	\end{matrix}\right. 
\end{align}
where $t$ represents the current training step, and $t_{warm}$ denotes the maximum warm training step.

\begin{table*}[h]
	\vspace{-1.2cm}
	\centering
	\caption{Quantitative comparison with eight state-of-the-art methods on the LA dataset. }
	\label{table2}
	\vspace{-0.2cm}
	\resizebox{0.9\textwidth}{!}{
		\begin{tabular}{llcclcccc}
			\hline
			\multirow{2}{*}{Method}                                 &                      & \multicolumn{2}{c}{Scans used}                           &                      & \multicolumn{4}{c}{Metrics}                                                                   \\ \cline{3-4} \cline{6-9} 
			&                      & Labeled                    & Unlabeled                   &                      & Dice$\uparrow$(\%) & Jaccard$\uparrow$(\%) & 95HD$\downarrow$(voxel) & ASD$\downarrow$(voxel) \\ \hline
			V-Net                                                   &                      & 4(5$\%$)                   & 0                           &                      & 52.55              & 39.60                 & 47.05                   & 9.87                   \\
			V-Net                                                   &                      & 8(10$\%$)                  & 0                           &                      & 78.57              & 66.96                 & 21.20                   & 6.07                   \\
			V-Net                                                   &                      & 80(100$\%$)                & 0                           &                      & 91.62              & 84.60                 & 5.40                    & 1.64                   \\ \hline
			DTC (AAAI'21) \cite{luo2021semi}       &                      & \multirow{9}{*}{4(5$\%$)}  & \multirow{9}{*}{76(95$\%$)} &                      & 81.25              & 69.33                 & 14.90                   & 3.99                   \\
			URPC (MedIA'22) \cite{luo2022semi}     &                      &                            &                             &                      & 82.48              & 71.35                 & 14.65                   & 3.65                   \\
			SS-Net (MICCAI'22) \cite{ssnet}                                                 &                      &                            &                             &                      & 86.33              & 76.15                 & 9.97                    & 2.31                   \\
			MC-Net+ (MedIA'22) \cite{wu2022mutual} &                      &                            &                             &                      & 82.07              & 70.38                 & 20.49                   & 5.72                   \\
			CAML (MICCAI'23) \cite{caml}           &                      &                            &                             &                      & 87.34              & 77.65                 & 9.76                    & 2.49                   \\
			Co-BioNet (Nat. Mach'23) \cite{peiris2023uncertainty}                                 &                      &                            &                             &                      & 84.30              & 74.67                 & 8.33                    & 2.38                   \\
			BCP (CVPR'23)\cite{Bai_2023_CVPR}    &                      &                            &                             &                      & 88.02              & 78.72                 & 7.90                    & 2.15                   \\
			MCF (CVPR'23)\cite{wang2023mcf}        &                      &                            &                             &                      & 83.34              & 72.20                 & 16.65                   & 5.46                   \\
			Ours                                                    &                      &                            &                             &                      & \textbf{89.70}              & \textbf{81.40}                 & \textbf{6.68}                    &\textbf{ 1.75 }                  \\ \hline
			DTC (AAAI'21) \cite{luo2021semi}       & \multicolumn{1}{c}{} & \multirow{9}{*}{8(10$\%$)} & \multirow{9}{*}{72(90$\%$)} & \multicolumn{1}{c}{} & 87.51              & 78.17                 & 8.23                    & 2.36                   \\
			URPC (MedIA'22) \cite{luo2022semi}     & \multicolumn{1}{c}{} &                            &                             & \multicolumn{1}{c}{} & 85.01              & 74.36                 & 15.37                   & 3.96                   \\
			SS-Net (MICCAI'22) \cite{ssnet}        & \multicolumn{1}{c}{} &                            &                             & \multicolumn{1}{c}{} & 88.43              & 79.43                 & 7.95                    & 2.55                   \\
			MC-Net+ (MedIA'22) \cite{wu2022mutual} & \multicolumn{1}{c}{} &                            &                             & \multicolumn{1}{c}{} & 88.96              & 80.25                 & 7.93                    & 1.86                   \\
			CAML (MICCAI'23) \cite{caml}           & \multicolumn{1}{c}{} &                            &                             & \multicolumn{1}{c}{} & 89.62              & 81.28                 & 8.76                    & 1.85                   \\
			Co-BioNet (Nat. Mach'23) \cite{peiris2023uncertainty}                                 & \multicolumn{1}{c}{} &                            &                             & \multicolumn{1}{c}{} & 89.20              & 80.68                 & 6.44                    & 1.90                   \\
			BCP (CVPR'23)\cite{Bai_2023_CVPR}    & \multicolumn{1}{c}{} &                            &                             & \multicolumn{1}{c}{} & 89.62              & 81.31                 & 6.81                    & \textbf{1.76}                   \\
			MCF (CVPR'23)\cite{wang2023mcf}        & \multicolumn{1}{c}{} &                            &                             & \multicolumn{1}{c}{} & 87.67              & 78.42                 & 9.16                    & 2.79                   \\
			Ours                                                    & \multicolumn{1}{c}{} &                            &                             & \multicolumn{1}{c}{} & \textbf{90.76}              & \textbf{83.13}                 & \textbf{6.08}                    & 1.87                   \\ \hline
	\end{tabular} }
\end{table*}
\section{Experience}

\vspace{-0.2cm}
\subsection{Dataset}
\subsubsection{LA dataset}
The LA dataset \cite{xiong2021global}, which serves as the benchmark dataset for the 2018 Atrial Segmentation Challenge, comprises 100 gadolinium-enhanced MR imaging scans for training, with a resolution of 0.625$\times$0.625$\times$0.625 mm. Due to the absence of publicly available annotations for the LA testing set, we allocated 80 samples for training and reserved the remaining 20 samples for validation following \cite{wu2022mutual}. For a fair comparison, we did not use any post-processing strategy.
\vspace{-0.2cm}
\subsubsection{Pancreas-CT dataset}
The Pancreas-CT dataset \cite{clark2013cancer} consists of 82 3D abdominal contrast-enhanced CT scans. The scans were obtained with resolutions of 512$\times$512 pixels and varying pixel sizes. For this study, we randomly selected 60 images for training and 20 images for testing, following a standard data splitting protocol commonly used in similar studies \cite{wu2022mutual}. To ensure consistency and comparability of voxel values, we applied a clipping operation to limit the values to the range of -125 to 275 Hounsfield Units (HU). 
\vspace{-0.2cm}
\subsubsection{BraTS2019 dataset}
The publicly available BraTS2019 dataset \cite{menze2014multimodal} includes scans obtained from 335 patients diagnosed with glioma. This dataset encompasses T1, T2, T1 contrast-enhanced, and FLAIR sequences, along with corresponding tumor segmentations annotated by expert radiologists. In this study, we focused on using the FLAIR modality for segmentation. We conducted a random split, allocating 250 scans for training, 25 scans for validation, and 60 scans for testing, as described in \cite{luo2022semi}.
\vspace{-0.3cm}
\subsection{Implementation Details}
\vspace{-0.1cm}
Our framework was developed using PyTorch and executed on an Nvidia RTX 3090 GPU equipped with 24GB of memory. 
\begin{figure*}[!ht]
	\centering
	\vspace{-0.3cm}
	\includegraphics[width=1.00\textwidth]{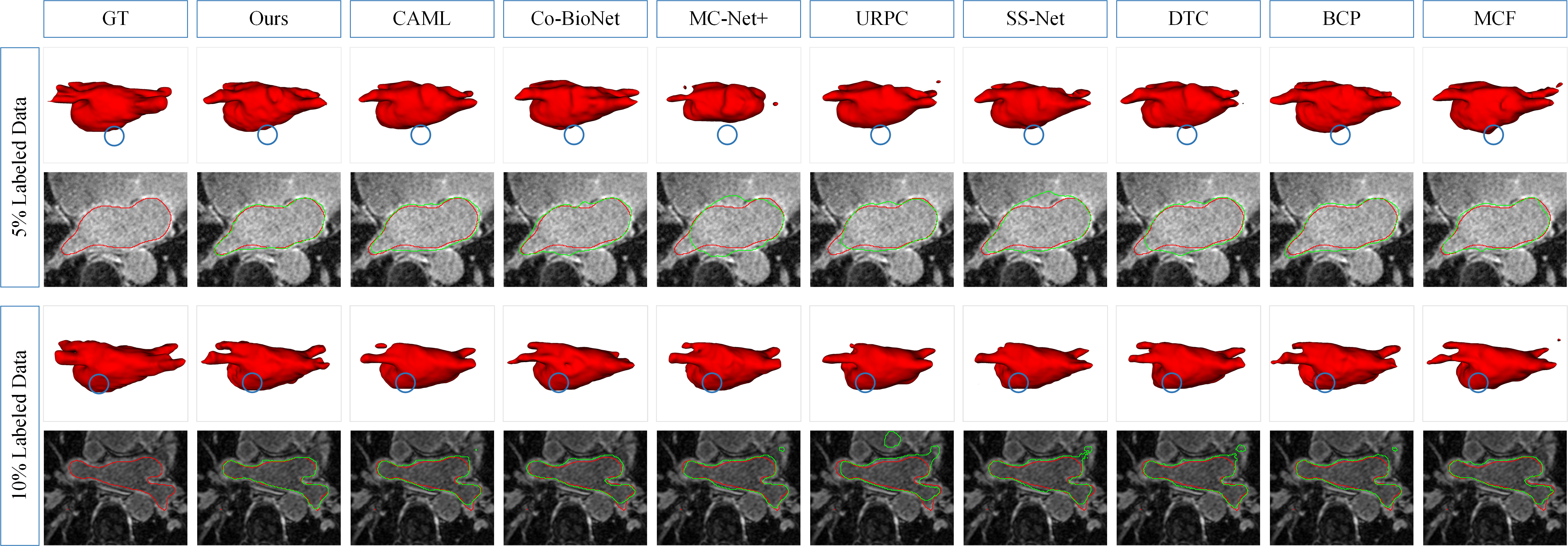}		
	\caption{Visualization of the segmentations results from different methods on the LA dataset. The red lines denote the boundary of GT and the green lines denote the boundary of predictions.}
	\label{figure2}
	\vspace{-0.5cm}
\end{figure*}
During training, we randomly extracted 3D patches from the preprocessed data. The patch size was set to 112 $\times$ 112 $\times$ 80 for the LA dataset and 96 $\times$ 96 $\times$ 96 for the Pancreas-CT and BraTS2019 datasets. The batch size was set to 4, comprising an equal distribution of two labeled and two unlabeled patches. We selected the V-Net \cite{milletari2016v} as the backbone of our framework to ensure a fair comparison with state-of-the-art (SOTA) method. The SGRS-Net framework was trained for 15$k$ iterations for the LA and Pancreas-CT dataset and 60$k$ iterations for the BraTS2019 dataset, using the SGD optimizer with a learning rate of 1e-2 and a weight decay factor 1e-4. The $\tau$, $\varepsilon$, and $\eta$ were set to 0.296, 0.2 and 20, respectively. Four commonly-used metrics are employed to assess the segmentation result: Dice, Jaccard, the average surface distance (ASD), and the 95$\%$ Hausdorff Distance (95HD). We compared our proposed framework with eight SOTA methods, including DTC, URPC, SS-Net, MC-Net+, CAML, Co-BioNet, BCP and MCF. Note that we utilized their official codes and results along with their publicly available data preprocessing schemes. 

\begin{table*}[h]
	\centering
	\caption{Quantitative comparison with eight SOTA methods on the Pancreas-CT dataset. }
	\label{table3}
	\vspace{-0.4cm}
	\resizebox{0.9\textwidth}{!}{
		\begin{tabular}{llcclcccc}
			\hline
			\multirow{2}{*}{Method}                                 &                      & \multicolumn{2}{c}{Scans used}                           &                      & \multicolumn{4}{c}{Metrics}                                                                   \\ \cline{3-4} \cline{6-9} 
			&                      & Labeled                    & Unlabeled                   &                      & Dice$\uparrow$(\%) & Jaccard$\uparrow$(\%) & 95HD$\downarrow$(voxel) & ASD$\downarrow$(voxel) \\ \hline
			V-Net                                                   &                      & 3(5$\%$)                   & 0                           &                      & 29.32              & 19.61                 & 43.67                   & 15.42                  \\
			V-Net                                                   &                      & 6(10$\%$)                  & 0                           &                      & 54.94              & 40.87                 & 47.48                   & 17.43                  \\
			V-Net                                                   &                      & 62(100$\%$)                & 0                           &                      & 83.76              & 72.48                 & 4.46                    & 1.07                   \\ \hline
			DTC (AAAI'21) \cite{luo2021semi}       &                      & \multirow{9}{*}{3(5$\%$)}  & \multirow{9}{*}{59(95$\%$)} &                      & 47.57              & 33.41                 & 44.17                   & 15.31                  \\
			URPC (MedIA'22) \cite{luo2022semi}     &                      &                            &                             &                      & 45.94              & 34.14                 & 48.80                   & \textbf{23.03}         \\
			SS-Net (MICCAI'22) \cite{ssnet}                                                 &                      &                            &                             &                      & 41.39              & 27.65                 & 52.12                   & 19.37                  \\
			MC-Net+ (MedIA'22) \cite{wu2022mutual} &                      &                            &                             &                      & 32.45              & 21.22                 & 58.57                   & 24.84                  \\
			CAML (MICCAI'23) \cite{caml}           &                      &                            &                             &                      & 35.18              & 23.63                 & 43.58                   & 20.39                  \\
			Co-BioNet (Nat. Mach'23) \cite{peiris2023uncertainty}                                 &                      &                            &                             &                      & 52.82              & 39.20                  & 29.46                   & 6.16                   \\
			BCP (CVPR'23)\cite{Bai_2023_CVPR}    &                      &                            &                             &                      & 45.08              & 34.72                 & 39.39                   & 11.23                  \\
			MCF (CVPR'23)\cite{wang2023mcf}        &                      &                            &                             &                      & 63.65              & 49.72                 & 18.06                   & 3.97                   \\
			Ours                                                    &                      &                            &                             &                      & \textbf{77.20}              & \textbf{63.45}                 & \textbf{12.34}                   & \textbf{3.81}                   \\ \hline
			DTC (AAAI'21) \cite{luo2021semi}       & \multicolumn{1}{c}{} & \multirow{9}{*}{6(10$\%$)} & \multirow{9}{*}{56(90$\%$)} & \multicolumn{1}{c}{} & 66.58              & 51.79                 & 15.46                   & 4.16                   \\
			URPC (MedIA'22) \cite{luo2022semi}     & \multicolumn{1}{c}{} &                            &                             & \multicolumn{1}{c}{} & 73.53              & 59.44                 & 22.57                   & 7.85                   \\
			SS-Net (MICCAI'22) \cite{ssnet}        & \multicolumn{1}{c}{} &                            &                             & \multicolumn{1}{c}{} & 73.44              & 58.82                 & 12.56                   & 2.91                   \\
			MC-Net+ (MedIA'22) \cite{wu2022mutual} & \multicolumn{1}{c}{} &                            &                             & \multicolumn{1}{c}{} & 70.00              & 55.66                 & 16.03                   & \textbf{3.87}          \\
			CAML (MICCAI'23) \cite{caml}           & \multicolumn{1}{c}{} &                            &                             & \multicolumn{1}{c}{} & 71.65              & 56.85                 & 14.87                   & 2.49                   \\
			Co-BioNet (Nat. Mach'23) \cite{peiris2023uncertainty}                                 & \multicolumn{1}{c}{} &                            &                             & \multicolumn{1}{c}{} & 77.89              & 64.79                 & 8.81                    & \textbf{1.39}                  \\
			BCP (CVPR'23)\cite{Bai_2023_CVPR}    & \multicolumn{1}{c}{} &                            &                             & \multicolumn{1}{c}{} & 73.49              & 58.60                 & 16.65                   & 2.22                   \\
			MCF (CVPR'23)\cite{wang2023mcf}        & \multicolumn{1}{c}{} &                            &                             & \multicolumn{1}{c}{} & 65.39              & 51.43                 & 13.85                   & 2.39                   \\
			Ours                                                    & \multicolumn{1}{c}{} &                            &                             & \multicolumn{1}{c}{} & \textbf{80.55}              & \textbf{67.88}                 & \textbf{6.00}                    & 2.50                   \\ \hline
	\end{tabular} }
\end{table*}
\vspace{-0.3cm}
\subsection{Result on the LA dataset}  
\vspace{-0.1cm}
Table \ref{table2} details the performance of our model compared to SOTA methods, alongside the fully supervised V-Net model at various ratios to delineate the lower and upper bounds of performance. Remarkably, our model achieved the best results across nearly all four metrics. Particularly impressive is the performance at only 5$\%$ labeled data, where our model achieved a Dice score of 89.70$\%$, marking a 2.36$\%$ improvement over the SOTA algorithm CAML (87.34$\%$). Even more striking is that with 5$\%$ labeled data, the Dice score of our model almost exceeds that of compared algorithms with 10$\%$ labeled data. With 10$\%$ labeled data, our approach attained a Dice score of 90.76$\%$, approaching the upper bounds. Figure \ref{figure2} offers a visual comparison of segmentation results from 2D and 3D perspectives alongside the corresponding GT. It clearly demonstrates that our SGRS-Net excels in segmentation accuracy, especially in 2D views, where our outcomes are notably closer to the GT than those of compared methods. These findings underscore the efficacy of our framework in the LA dataset.
\begin{figure*}[!ht]
	\centering
	\vspace{-0.3cm}
	\includegraphics[width=1.00\textwidth]{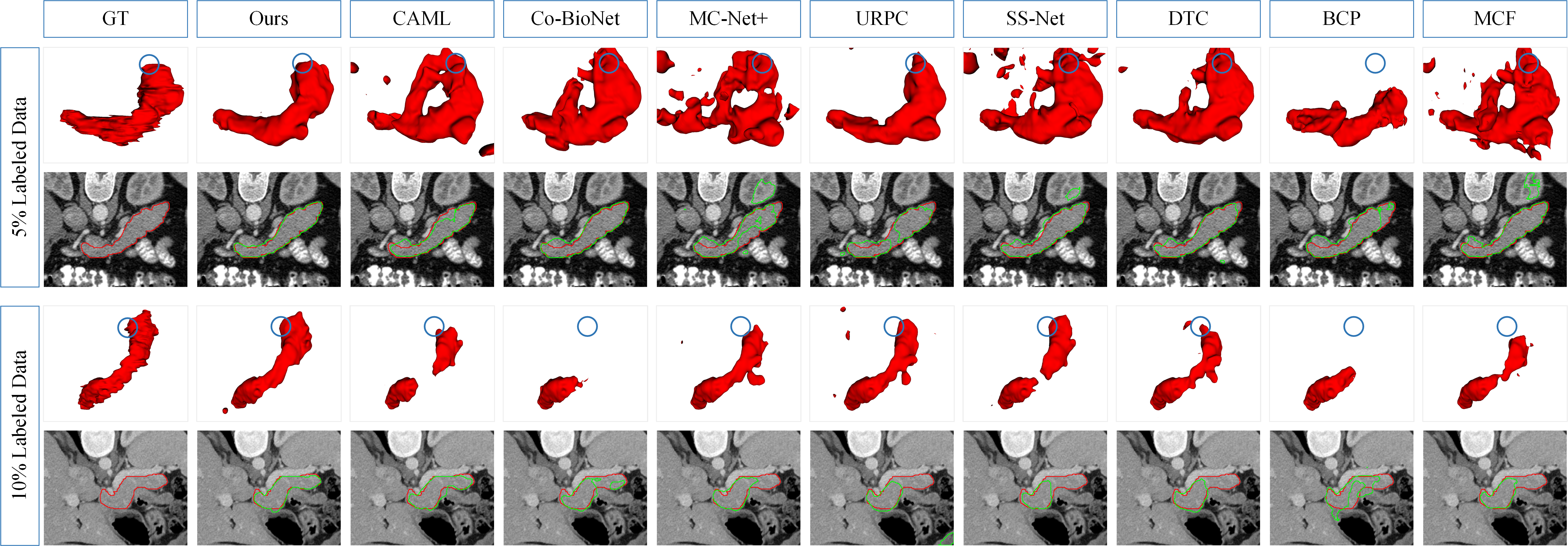}		
	\caption{Visualization of the segmentations results from different methods on the Pancreas-CT dataset. The red lines denote the boundary of GT and the green lines denote the boundary of predictions.}
	\label{figure3}
		\vspace{-0.5cm}
\end{figure*}
\vspace{-0.2cm}
\subsection{Result on the Pancreas-CT dataset}
Table \ref{table3} presents a quantitative comparison on the Pancreas-CT dataset. The results indicate that our model stands out prominently. With 5$\%$ labeled data, we achieved a Dice score of 77.20$\%$, a 95HD of 12.34 and an ASD of 3.81. Which is a 9.55$\%$ improvememt over the SOTA method MCF(63.65$\%$). These metrics not only surpass the accuracy obtained by the compared models based on equivalent data annotation, but also match or surpass the the performance of these models with 10$\%$ labeled data. Figure \ref{figure3} illustrates the segmentation results on the Pancreas-CT dataset. It is apparent that with 5$\%$ labeled data, the compared models fail to segment an approximate pancreas, while our model achieves segmentation results relatively closer to the GT. In the 2D visual representation, the compared method exhibits a higher rate of false negatives, whereas our model achieves more precise identification. 
These further underscore the effectiveness of the proposed approach.
\vspace{-0.2cm}
\subsection{Results on the BraTS2019 dataset}
Table \ref{table4} presents the quantitative experimental results for the BraTS2019 dataset. Compared to SOTA models, our model achieved improved performance Dice scores. Specificly, we achieved a Dice score of 80.62$\%$ using 5$\%$ labeled data and 85.67$\%$ using $10\%$ labeled data. Which are 1.28$\%$ and 1.32$\%$ improvement over the SOTA method CAML(79.34$\%$) and BCP(84.35$\%$). These further emphasize the effectiveness of the proposed model.

\begin{table*}[h]
	\centering
	\caption{Quantitative comparison with eight SOTA methods on the BraTS2019 dataset. }
	\label{table4}
	\vspace{-0.3cm}
	\resizebox{0.9\textwidth}{!}{
		\begin{tabular}{llcclcccc}
			\hline
			\multirow{2}{*}{Method}                                 &                      & \multicolumn{2}{c}{Scans used}                             &                      & \multicolumn{4}{c}{Metrics}                                                                   \\ \cline{3-4} \cline{6-9} 
			&                      & Labeled                     & Unlabeled                    &                      & Dice$\uparrow$(\%) & Jaccard$\uparrow$(\%) & 95HD$\downarrow$(voxel) & ASD$\downarrow$(voxel) \\ \hline
			V-Net                                                   &                      & 12(5$\%$)                   & 0                            &                      & 74.28              & 64.42                 & 13.44                   & 2.60                   \\
			V-Net                                                   &                      & 25(10$\%$)                  & 0                            &                      & 78.67              & 68.75                 & 10.44                   & 2.23                   \\
			V-Net                                                   &                      & 250(100$\%$)                & 0                            &                      & 88.58              & 80.34                 & 6.19                    & 1.36                   \\ \hline
			DTC (AAAI'21) \cite{luo2021semi}       &                      & \multirow{9}{*}{12(5$\%$)}  & \multirow{9}{*}{238(95$\%$)} &                      & 74.21              & 64.89                 & 13.54                   & 3.16                   \\
			URPC (MedIA'22) \cite{luo2022semi}     &                      &                             &                              &                      & 78.74              & 68.20                 & 17.43                   & 4.51          \\
			SS-Net (MICCAI'22) \cite{ssnet}                                                  &                      &                             &                              &                      & 78.03              & 68.11                 & 13.70                   & 2.76                   \\
			MC-Net+ (MedIA'22) \cite{wu2022mutual} &                      &                             &                              &                      & 78.69              & 68.38                 & 16.44                   & 4.49                   \\
			CAML (MICCAI'23) \cite{caml}           &                      &                             &                              &                      & 79.34              & 69.64                 & \textbf{11.02}                   & 2.36                   \\
			Co-BioNet (Nat. Mach'23) \cite{peiris2023uncertainty}                                 &                      &                             &                              &                      & 73.27              & 63.12                 & 14.37                   & \textbf{1.90}                   \\
			BCP (CVPR'23)\cite{Bai_2023_CVPR}    &                      &                             &                              &                      & 79.30              & 68.89                 & 12.00                   & 1.94                   \\
			MCF (CVPR'23)\cite{wang2023mcf}        &                      &                             &                              &                      & 71.82              & 62.60                 & 14.66                   & 3.96                   \\
			Ours                                                    &                      &                             &                              &                      & \textbf{80.62}              & \textbf{70.21}                 & 14.58                   & 3.64                   \\ \hline
			DTC (AAAI'21) \cite{luo2021semi}       & \multicolumn{1}{c}{} & \multirow{9}{*}{25(10$\%$)} & \multirow{9}{*}{225(90$\%$)} & \multicolumn{1}{c}{} & 82.74              & 72.74                 & 11.76                   & 3.24                   \\
			URPC (MedIA'22) \cite{luo2022semi}     & \multicolumn{1}{c}{} &                             &                              & \multicolumn{1}{c}{} & 84.16              & 74.29                 & 11.01                   & 2.63                   \\
			SS-Net (MICCAI'22) \cite{ssnet}        & \multicolumn{1}{c}{} &                             &                              & \multicolumn{1}{c}{} & 82.00              & 71.82                 & 10.68                   & \textbf{1.82}                   \\
			MC-Net+ (MedIA'22) \cite{wu2022mutual} & \multicolumn{1}{c}{} &                             &                              & \multicolumn{1}{c}{} & 79.63              & 70.10                 & 12.28                   & 2.45         \\
			CAML (MICCAI'23) \cite{caml}           & \multicolumn{1}{c}{} &                             &                              & \multicolumn{1}{c}{} & 81.58              & 72.31                 & 10.30                   & 1.94                   \\
			Co-BioNet (Nat. Mach'23) \cite{peiris2023uncertainty}                                 & \multicolumn{1}{c}{} &                             &                              & \multicolumn{1}{c}{} & 75.22              & 65.32                 & 13.56                   & 1.94                   \\
			BCP (CVPR'23)\cite{Bai_2023_CVPR}    & \multicolumn{1}{c}{} &                             &                              & \multicolumn{1}{c}{} & 84.35              & 75.01                 & 10.95                   & 2.60                   \\
			MCF (CVPR'23)\cite{wang2023mcf}        & \multicolumn{1}{c}{} &                             &                              & \multicolumn{1}{c}{} & 79.28              & 69.08                 & 12.43                   & 3.58                   \\
			Ours                                                    & \multicolumn{1}{c}{} &                             &                              & \multicolumn{1}{c}{} & \textbf{85.67}              & \textbf{76.05}                 & \textbf{9.33}                    & 2.76                  \\ \hline
	\end{tabular}  }
\end{table*}

\begin{table*}[h]
	\centering
	\caption{Ablation study on the components of the proposed SGRS-Net. The "MT" refers to the integration with the mean-teacher framework, which converts the teacher network's predictions into pseudo labels.}
	\label{tabe5}
	\vspace{-0.2cm}
	\resizebox{0.9\textwidth}{!}{
		\begin{tabular}{cccccccccccc}
			\hline
			\multicolumn{4}{c}{Components}                             &  & \multicolumn{2}{c}{Scans used}           &  & \multicolumn{4}{c}{Metrics}                       \\ \cline{1-4} \cline{6-7} \cline{9-12} 
			Baselibe & MA & MT & SE$\&$RLE &  & Labeled            & Unlabeled           &  & Dice$\uparrow$(\%) & Jaccard$\uparrow$(\%) & 95HD$\downarrow$(voxel) & ASD$\downarrow$(voxel) \\ \hline
			\checkmark      &    &             &              &  & \multirow{5}{*}{4(5$\%$)} & \multirow{5}{*}{76(95$\%$)} &  & 75.18    & 63.6        & 22.74       & 4.97       \\
			\checkmark      & \checkmark      &             &              &  &                    &                     &  & 80.43    & 69.66       & 18.94       & 4.97       \\
			\checkmark      &             & \checkmark      &              &  &                    &                     &  & 87.45    & 77.95       & 8.95        & \textbf{2.09}       \\
			\checkmark      & \checkmark      & \checkmark      &              &  &                    &                     &  & 88.61    & 79.65       & 11.09       & 2.84       \\
			\checkmark      & \checkmark      & \checkmark      & \checkmark       &  &                    &                     &  &\textbf{89.72}    & \textbf{81.42}       &\textbf{ 8.46 }       & 2.42       \\ \hline
			\checkmark      & \textbf{}   &             &              &  & \multirow{5}{*}{8(10$\%$)} & \multirow{5}{*}{(90$\%$)} &  & 81.82    & 70.01       & 26.64       & 7.25       \\
			\checkmark      & \checkmark      &             &              &  &                    &                     &  & 85.87    & 75.45       & 12.33       & 3.11       \\
			\checkmark      &             & \checkmark      &              &  &                    &                     &  & 89.66    & 81.35       & 7.59        & 2.21       \\
			\checkmark      & \checkmark      & \checkmark      &              &  &                    &                     &  & 89.90    & 81.72       & 6.15        & 1.92       \\
			\checkmark      & \checkmark      & \checkmark      & \checkmark       &  &                    &                     &  & \textbf{90.76}    & \textbf{83.13}       & \textbf{6.08}       & \textbf{1.87}       \\ \hline
	\end{tabular}	}
	\vspace{-0.2cm}
\end{table*}

\begin{figure*}[!h]
	\begin{minipage}[h]{0.58\textwidth}
		
		\captionsetup{type=table}
		\caption{Effect of the loss function for different regions.}
		\label{tabe6}
		\vspace{-0.4cm}
		\resizebox{1\textwidth}{!}{
			\begin{tabular}{ccclcclcccc}
				\hline
				\multicolumn{3}{c}{Regions} &  & \multicolumn{2}{c}{Scans used}           &  & \multicolumn{4}{c}{Metrics}                       \\ \cline{1-3} \cline{5-6} \cline{8-11} 
				$\mathrm{\Omega}$  & $\mathrm{\Theta}$ & $\mathrm{\Delta}$ &  & Labeled            & Unlabeled           &  & Dice$\uparrow$(\%) & Jaccard$\uparrow$(\%) & 95HD$\downarrow$(voxel) & ASD$\downarrow$(voxel) \\ \hline
				$\mathcal{L}_{con}$    & $\mathcal{L}_{con}$     &       &  & \multirow{6}{*}{4(5$\%$)} & \multirow{6}{*}{76(95$\%$)} &  & 89.04    & 80.32       & 7.45        & 1.97       \\
				$\mathcal{L}_{NR}$     & $\mathcal{L}_{con}$     &       &  &                    &                     &  & 88.14    & 79.02       & 12.95       & 2.81       \\
				$\mathcal{L}_{NR}$     & $\mathcal{L}_{NR}$      &       &  &                    &                     &  & 88.76    & 79.96       & 7.39        & 1.96       \\
				$\mathcal{L}_{NR}$     & $\mathcal{L}_{NR}$      & $\mathcal{L}_{NR}$    &  &                    &                     &  & 88.61    & 79.67       & 8.52        & 2.21       \\
				$\mathcal{L}_{con}$    & $\mathcal{L}_{con}$     & $\mathcal{L}_{con}$   &  &                    &                     &  & 88.61    & 79.65       & 11.09       & 2.84       \\
				$\mathcal{L}_{con}$    & $\mathcal{L}_{NR}$      &       &  &                    &                     &  & \textbf{89.70}    & \textbf{81.40}       & \textbf{6.68}        & \textbf{1.75}       \\ \hline
				$\mathcal{L}_{con}$    & $\mathcal{L}_{con}$     &       &  & \multirow{6}{*}{8(10$\%$)} & \multirow{6}{*}{72(90$\%$)} &  & 90.05    & 81.97       & 6.45        & 2.11       \\
				$\mathcal{L}_{NR}$     & $\mathcal{L}_{con}$     &       &  &                    &                     &  & 90.07    & 82.02       & 7.06        & 2.20        \\
				$\mathcal{L}_{NR}$     & $\mathcal{L}_{NR}$      &       &  &                    &                     &  & 90.12    & 82.10        & 6.82        & 2.04       \\
				$\mathcal{L}_{NR}$     & $\mathcal{L}_{NR}$      & $\mathcal{L}_{NR}$    &  &                    &                     &  & 89.82    & 81.65       & 7.07        & 2.01       \\
				$\mathcal{L}_{con}$    & $\mathcal{L}_{con}$     & $\mathcal{L}_{con}$   &  &                    &                     &  & 89.90    & 81.72       & 6.15        & 1.92       \\
				$\mathcal{L}_{con}$    & $\mathcal{L}_{NR}$      &       &  &                    &                     &  & \textbf{90.76}    & \textbf{83.13}       & \textbf{6.08}        & \textbf{1.87}       \\ \hline
		\end{tabular}}
		
	
		\vspace{0.2 cm}
		\captionsetup{type=table}
		\caption{Effect of the loss function for different regions.}
		\label{tabel7}
		\vspace{-0.4cm}
		\resizebox{1\textwidth}{!}{
			\begin{tabular}{llcclcccc}
				\hline
				\multicolumn{1}{c}{\multirow{2}{*}{Method}} &  & \multicolumn{2}{c}{Scans used}           &  & \multicolumn{4}{c}{Metrics}                       \\ \cline{3-4} \cline{6-9} 
				\multicolumn{1}{c}{}                        &  & Labeled            & Unlabeled           &  & Dice$\uparrow$(\%) & Jaccard$\uparrow$(\%) & 95HD$\downarrow$(voxel) & ASD$\downarrow$(voxel) \\ \hline
				Flip.H                                     &  & \multirow{3}{*}{4(5$\%$)} & \multirow{3}{*}{76(95$\%$)} &  & 88.69    & 79.75       & 10.39        & 2.50       \\
				Flip.V                                     &  &                    &                     &  & 88.96    & 80.21       & 7.57        & 2.18       \\
				MA                                          &  &                    &                     &  & \textbf{89.70}    & \textbf{81.40}       &\textbf{6.68}        & \textbf{1.75}       \\ \hline
				Flip.H                                     &  & \multirow{3}{*}{8(10$\%$)} & \multirow{3}{*}{72(90$\%$)} &  & 90.06    & 81.98       & 6.91        & 1.96       \\
				Flip.V                                     &  &                    &                     &  & 90.31    & 82.43       & 6.12        & 1.99       \\
				MA                                          &  &                    &                     &  & \textbf{90.76}    & \textbf{83.13}       & \textbf{6.08}        & \textbf{1.87}       \\ \hline
		\end{tabular}}
	\end{minipage}
	\hfill
	\begin{minipage}[h]{0.38\textwidth}
		\includegraphics[width=1\textwidth]{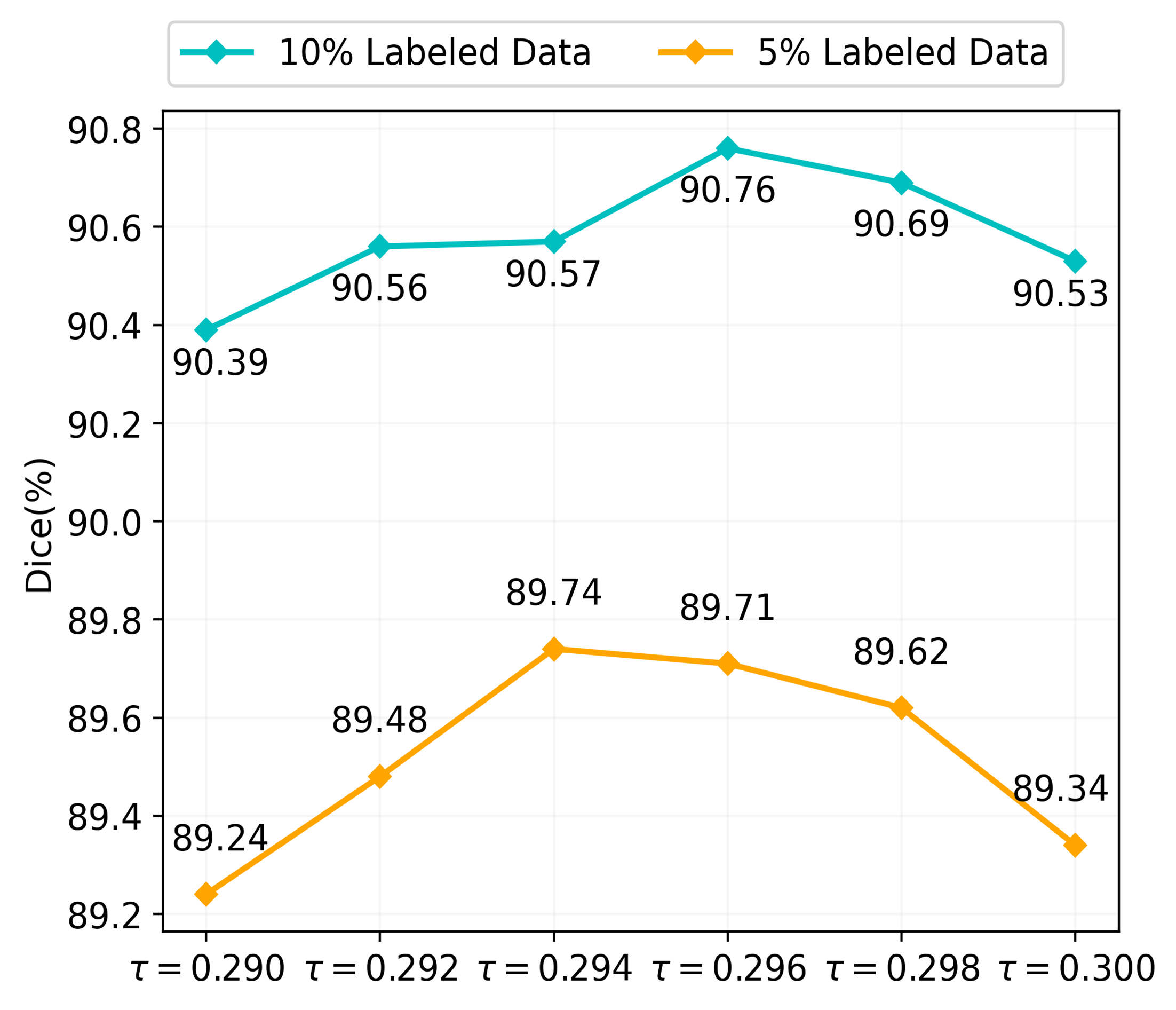}		
		\captionsetup{type=figure}
		\caption{Illustrations of corresponding Dice score with different $\tau$ value on the LA dataset.}
		\label{figure5}
	\end{minipage}
	\vspace{-0.6cm}
\end{figure*}
\vspace{-0.3cm}
\section{Ablation study}
\vspace{-0.2cm}
\subsection{Effect of the components}
\vspace{-0.1cm}
We perform an ablation study to evaluate the effects of the MA module, the combination with the Mean-teacher framework, and the SE$\&$RLE module. We use the pseudo labels generated by single V-Net as the baseline. The results are detailed in Table \ref{tabe5}. These results underscore that the integrating pseudo labels with the Mean-Teacher framework leads to a marked enhancement in performance. Moreover, integrating the SE$\&$RLE module to utilize pseudo labels separately leads to further improvements. With 10$\%$ labeled data, the Dice score progresses from 89.90$\%$ to 90.76$\%$. Notably, this is a 0.86$\%$ improvement that approaching the upper bound. This demonstrates the effective use of pseudo labels by the SEM$\&$RLE modules and underscores their crucial role in enhancing model accuracy.
\vspace{-0.3cm}
\subsection{Effect of the loss functions for distinct regions in $\mathcal{Y}$}
\vspace{-0.1cm}
In this paper, we used $\mathcal{L}_{con}$ and $\mathcal{L}_{NR}$ to evaluate the losses within $\mathrm{\Omega}$ and $\mathrm{\Theta}$, while excluding the evaluation of regions within $\mathrm{\Delta}$. To validate the rationality of the RLE module, we conducted ablation studies on the loss functions for each region, as shown in Table \ref{tabe6}. The results indicate that excluding the $\mathrm{\Delta}$ region enhances the model's robustness. Traditional methods that used $\mathcal{L}_{con}$ for the entire $\mathcal{Y}$ achieved Dice score of 88.61$\%$ with 5$\%$ labeled data. In contrast, our method achieved better performance with Dice score of 89.70$\%$, underscoring the effectiveness of the RLE module.

\vspace{-0.2cm}
\subsection{Effect of the augumentation method}
In this study, we present the MA module, which uses $\mathcal{D_{L}}$ to augment $\mathcal{D_{U}}$, forming the foundation for the proposed RLE and SEM modules. Various data augmentation methods, such as the commonly used horizontal and vertical flipping are also applicable to our framework. We compared the effects of horizontal flipping, vertical flipping, and the MA module, with the results presented in Table \ref{tabel7}. These results show that flip-based augmentations also perform well. With 5$\%$ labeled data, Flip.H achieved a Dice score of 88.69$\%$, surpassing the SOTA method CAML (87.34$\%$). Moreover, the MA module achieved the best performance, prompting us to use it for augmenting and perturbing unlabeled data.
\vspace{-0.3cm}
\subsection{Effect of the parameter $\tau$}
\vspace{-0.1cm}
This study introduces the SEM to categorizes pseudo-labels into disregarded and considered sections sections using the parameter $\tau$. Figure \ref{figure5} illustrates the results on the LA dataset with various $\tau$ values. Smaller $\tau$ values result in a larger number of areas being disregarded, thus reducing supervision. Conversely, larger $\tau$ values can increase uncertainty in pseudo-labels, increasing the risk of noise. The results demonstrate that the overall performance of the model is better when $\tau$=0.296. Therefore, we chose $\tau$=0.296 for the subsequent experiments. 
\vspace{-0.2cm}
\section{Conclusion}
In this paper, we introduce SGRS-Net. Built upon the mean teacher network, we employ the MA module to enhance unlabeled data. By analyzing the synergy between predictions before and after data augmentation, we partition pseudo labels into distinct regions. Additionally, we present a region-based loss evaluation module to assess the loss within each identified area. Extensive experiments conducted on LA, Pancreas-CT and BraTS2019 datasets show the superiority of the proposed SGRS-Net, with even over 9.55$\%$ Dice improvement on Pancreas-CT dataset with 5$\%$ labeled data. These results underscore the efficiency and practicality of our framework.

\bibliography{reference_ssl}

\end{document}